%% file: main.tex
\pdfoutput=1
\documentclass{article}

\PassOptionsToPackage{numbers,sort&compress}{natbib}
\usepackage[preprint]{neurips_2025}

\usepackage[utf8]{inputenc}
\usepackage[T1]{fontenc}
\usepackage{url}
\usepackage{booktabs}
\usepackage{amsmath}
\usepackage{amssymb}
\usepackage{mathtools}
\usepackage{bm}
\usepackage{graphicx}
\usepackage{xcolor}
\usepackage{etoolbox}
\usepackage{enumitem}
\usepackage{float}
\usepackage{microtype}
\usepackage[colorlinks,allcolors=blue!60!black]{hyperref}
\usepackage[capitalise]{cleveref}

\input{sections/macros}
\makeatletter
\providecommand{\@LN}[2]{}
\providecommand{\@LN@col}[1]{}
\makeatother

\title{A Computational Model of Continuous Recognition Memory for Natural Images}

\author{%
  Sheng Zhao \\
  University of Rochester \\
  \texttt{szhao32@ur.rochester.edu}
  \And
  Yijin Wang \\
  American University \\
  \texttt{yw1695a@american.edu}
  \AND
  Bei Xiao \\
  American University \\
  \texttt{bxiao@american.edu}
  \And
  Yuhao Zhu \\
  University of Rochester \\
  \texttt{yzhu@rochester.edu}
}

\begin{document}
\maketitle
\suppressfloats[t]
\input{sections/abstract}
\input{sections/introduction}
\input{sections/related}
\input{sections/method}
\input{sections/experiments}
\input{sections/discussion}
\input{sections/conclusion}

{
\small
\bibliographystyle{ieeenat_fullname}
\bibliography{references}
}

\clearpage
\appendix
\raggedbottom
\input{sections/appendix}

\end{document}

%% file: sections/macros.tex
\newcommand{\R}{\mathbb{R}}

\def\cN{{\mathcal{N}}}

\renewcommand{\paragraph}[1]{\textbf{#1}~~}

\newcommand{\Sect}[1]{\Cref{#1}}
\newcommand{\Fig}[1]{\Cref{#1}}
\newcommand{\Tbl}[1]{\Cref{#1}}
\newcommand{\Equ}[1]{\cref{#1}}
\newcommand{\Apx}[1]{\Cref{#1}}
\AtEndPreamble{%
  \crefname{equation}{Eq.}{Eqs.}
  \Crefname{equation}{Eq.}{Eqs.}
  \creflabelformat{equation}{#2#1#3}
  \crefname{figure}{Fig.}{Figs.}
  \Crefname{figure}{Figure}{Figures}
  \crefname{appendix}{Appendix}{Appendices}
  \Crefname{appendix}{Appendix}{Appendices}
}

\newcommand{\RNum}[1]{\uppercase\expandafter{\romannumeral #1\relax}}

%% file: sections/abstract.tex
\begin{abstract}
Recognizing an image we have seen before is a basic function of human visual memory.
Understanding human continuous recognition memory (CRM) can help machine vision systems predict which images people are likely to remember and generate more memorable images.
It can also inform the diagnosis of memory impairments and the development of interventions to improve memory.
This paper proposes the first computational model for CRM.
The model captures three key components of continuous recognition memory (CRM): image encoding, memory decay, and decision making.
It encodes each image as a distinct random variable in a perceptual space.
As the memory of an image fades, the distribution of its random variable drifts within that space.
To decide whether the current image has been seen before, the observer retrieves samples from the distributions associated with previously encoded images and compares them with the current image.
Our model predicts human observers’ trial-by-trial responses in CRM tasks within the variability observed across human participants.
Our model is parameter efficient, requiring only four shared parameters and two parameters per observer.
It can accurately predict the memorability of an image no worse than prior methods that require millions of parameters.
\end{abstract}

%% file: sections/introduction.tex
\section{Introduction}
\label{sec:intro}

\begin{figure*}[t]
\centering
\includegraphics[width=\textwidth]{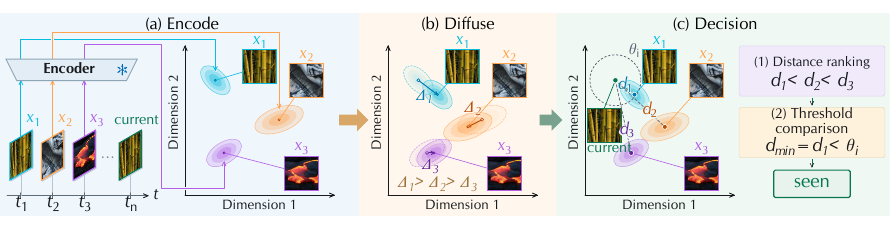}
\caption{\textbf{Overview of the observer model.}
(a): Encode: each image is represented as an image-specific distribution in a low-dimensional perceptual space, whose variance is larger for images in denser regions and for images with a weaker IT-aligned response.
(b): Diffuse: over time, the memory of each image drifts in a random direction within the low-dimensional space.
(c): Decide: a sample of the current image is compared with samples from the diffused memories of all earlier images.
The observer reports ``seen'' if the nearest distance is below the observer's decision threshold.
Repeated sampling under different random drifts gives the probability of this response.
}
\label{fig:observer}
\vspace{-3pt}
\end{figure*}

People can recognize thousands of images that they have seen only once~\cite{standing1973learning, brady2008visual}.
Predicting what people remember has inspired a body of work in computer vision, from large-scale datasets~\cite{khosla2015understanding, goetschalckx2019memcat} and deep networks that predict the memorability of images~\cite{khosla2015understanding, needell2022embracing} to methods that edit images to make them more memorable~\cite{goetschalckx2019ganalyze}.

Fundamentally, however, existing work has focused on memorability as a property of \textit{images}, with less emphasis on the memory processes of \textit{human} observers:
the mechanisms underlying continuous recognition memory (CRM), a fundamental aspect of memory that allows us to recognize previously seen images as we encounter new ones, remain poorly understood. 
Understanding these mechanisms can improve predictions of image memorability and, more importantly, help identify deficits in how people encode, store, and retrieve visual information. Identifying where these deficits arise could inform the diagnosis of memory impairments and the development of targeted interventions~\cite{ashford2019computerized}.

We propose the first computational model of the CRM.
We focus on a classical CRM paradigm used by most memorability datasets, where a participant views a sequence of images and reports whether an image has been seen before. 
Our model formalizes three assumptions underlying CRM.

First, visual encoding is noisy, so the model represents each image as a random variable in a perceptual space~\cite{zhao2026multidimensional}.
Retrieving an image from memory amounts to drawing a sample from its distribution.
We construct the perceptual space using frozen vision encoders that span multiple levels of the visual hierarchy: a retinal contrast model~\cite{peli1990contrast}, a model of cortical processing in the ventral visual stream~\cite{kubilius2019brain}, and CLIP for higher-level semantic representations~\cite{radford2021learning}.

Second, as the memory of an image fades, its distribution drifts and diffuses in the perceptual space.
Following classical work in cognitive neuroscience~\cite{compte2000synaptic, wimmer2014bump}, we model the drift as a Brownian motion.
Images that were seen a long time ago thus are given more time to drift and, on expectation, are more distant from their initial location in the perceptual space.

Third, to decide whether the current image has been seen before, the observer compares (a sample of) the current image with one sample retrieved for each previously presented image.
The observer reports having seen the current image if at least one retrieved sample lies sufficiently close, within a threshold, to the current image in perceptual space.

The explicit assumptions allow our model to be interpretable and extremely parameter efficient.
We fit the model to human data gathered from CRM experiments using maximum likelihood estimation.

Across four datasets where the interval between the two appearances of an image ranges from seconds to about 300 days~\cite{wang2026material, goetschalckx2019memcat, allen2022massive, kramer2023features}, our model achieves state-of-the-art accuracy in predicting trial-by-trial responses, outperforming LSTMs with orders of magnitude fewer parameters.
Its predictions are within the range of variability observed across human observers. 
While trained without supervision on image memorability statistics, our model also matches or exceeds the accuracy of prior memorability predictors~\cite{needell2022embracing} with orders of magnitude fewer parameters.

Our ablation analyses show that accurate trial-by-trial predictions depend on modeling the image distribution variance using two factors: image distinctiveness and the absolute magnitude of inferior temporal (IT) cortex to the image.
Image distinctiveness captures how distinct an image is from the other images presented, and the IT area represents high-level visual information such as object shape and identity.
Our findings suggest that recognition memory depends on both the context in which an image appears and how strongly the image engages object-selective representations in the visual system.

Our model also identifies two potential sources of individual differences in CRM performance. 
Observers differ in both the diffusion constant that governs memory decay and the threshold they use to make recognition decisions.
Variation in these parameters accounts for 59\% and 13\% of the observed differences in performance, respectively.

This paper makes the following contributions:
\begin{itemize}[leftmargin=1.5em, itemsep=2pt, parsep=0pt]
  \item We propose a computational observer model of continuous recognition memory that represents each image as a distribution in a low-dimensional perceptual space, where the distributions drift as memory decays and memory recognition depends on similarity judgment in the perceptual space.
  \item Our model predicts both human behaviors and image memorability statistics accurately.
  With few interpretable parameters, our model achieves state-of-the-art accuracy in predicting trial-by-trial human responses across four commonly used datasets.
  Without further tuning, it also rivals deep networks trained specifically to predict image memorability that use orders of magnitude more parameters.

  \item We show that the image-specific variance encoding explains between-image variations in recognition, and that the distinctiveness of an image and its IT-layer response account for much of this variance.
  Each person's decision threshold and diffusion constant well explain individual differences in recognition memory.
\end{itemize}

%% file: sections/related.tex
\section{Related work}
\label{sec:related}

\paragraph{Image Memorability.}
Since \citet{isola2011makes} showed that people agree on which photographs they remember, a body of work has measured and predicted image memorability, defined for an image as the likelihood that any given person will remember that image later.
Datasets such as LaMem~\cite{khosla2015understanding}, FIGRIM~\cite{bylinskii2015intrinsic}, and MemCat~\cite{goetschalckx2019memcat} measure memorability with online memory games, and deep networks such as MemNet~\cite{khosla2015understanding} and ResMem~\cite{needell2022embracing} regress memorability scores from natural images using pretrained vision encoders~\cite{he2016deep}.
Other work studies what makes an image memorable~\cite{isola2011makes, jaegle2019population, kramer2023features, newman2020multimodal, goetschalckx2019ganalyze}, but these methods can not process single trial.

\paragraph{Models of Recognition Memory.}
Psychological models of recognition memory explain single decisions with a few interpretable mechanisms.
They include signal detection theory~\cite{green1966signal}, global matching~\cite{gillund1984retrieval, hintzman1988judgments, shiffrin1997model}, and exemplar models~\cite{nosofsky1991tests}, and recent work applies them to images~\cite{nosofsky2026mds, schurgin2020psychophysical, annis2021combining, bates2024scaling, simpson2026memorability}.
However, these models are typically fit to aggregate data, such as memorability scores.
They also focus on simple designed stimuli or very small image sets, rather than large datasets of natural images.
Observer model is also used to interpret human perception doing image quality assessment, which represents images as distributions in a perceptual space~\cite{zhao2026multidimensional}.
We extend this idea to memory and fit it to single trials.

%% file: sections/method.tex
\section{Method}
\label{sec:method}
\subsection{Recognition Memory Task}
\label{sec:task}

\begin{figure*}[t]
\centering
\includegraphics[width=\textwidth]{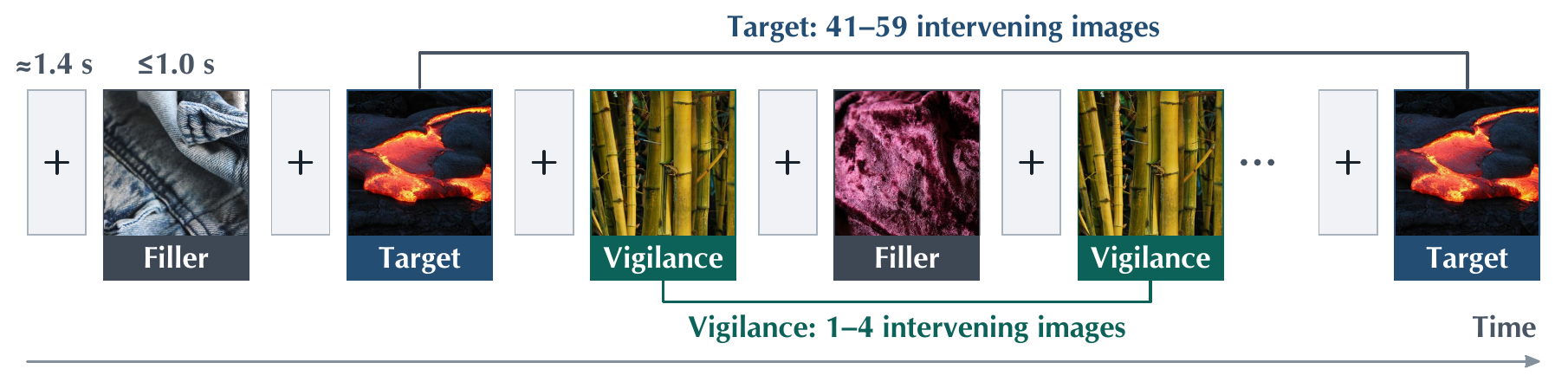}
\caption{Continuous recognition task, shown for the Material dataset~\cite{wang2026material} with images from the STUFF dataset~\cite{schmidt2025core}.
Each trial shows a fixation cross for about 1.4\,s and then an image for up to 1\,s, during which the observer presses a key if the image has appeared before.
A target image repeats after 41--59 intervening images, a vigilance image repeats after 1--4 intervening images to check attention, and a filler image appears once.}
\label{fig:paradigm}
\end{figure*}

Most prior work probes recognition memory with a continuous recognition task~\cite{isola2011makes, khosla2015understanding, goetschalckx2019memcat}.
\Fig{fig:paradigm} shows such a sequence from the Stuff dataset~\cite{wang2026material, schmidt2025core}, which includes images of materials.
An observer views a long sequence of images, each shown for half a second to a few seconds, and some images, which we call target images, appear again between 41 and 59 intervening images\footnote{This time gap varies significantly, from seconds to days, across different datasets.
As we will show later, our model can reliably predict across different times scales.}.
Other filler images will appear only once.
On each trial, the observer reports whether the current image has appeared before by pressing a button to indicate ``seen''~\cite{allen2022massive}.
A ``seen'' response to a repeated image is a hit and to a new image is a false alarm.
Throughout a session, some images reappears after only 1 to 4 intervening images;
these are ``vigilance'' tests to ensure that the participants stay engaged.

Our main goal is to predict every single response, which depends on the current image and on the images that this observer has seen before.
We will later show that our model can, without further tuning, also predict memorability score per-image.

\subsection{Observer Model}
\label{sec:observer}
To understand how people recognize images from behavioral data, we must make assumptions about how observers encode images, how their memories decay, and how they decide whether an image has been seen.
These assumptions are formalized in our \textit{observer model}.
Concretely, an observer's visual system encodes each image with a noisy pattern of neural activity, and memory representations decay over time~\cite{wixted1991form}.
Observers recognize a previously seen image by comparing the neural encoding of the incoming image against representations retrieved from memory. 
Because both the encoding and memories are noisy, the same observer can give different answers to the same image.

Our observer model formalizes these intuitions in three steps, shown in \Fig{fig:observer}: encoding, memory decay, and decision.

First, we represent each image $x$ in an $N$-dimensional perceptual space as a Gaussian random variable~\cite{zhao2026multidimensional} (\Fig{fig:observer}(a)):
\begin{equation}
  \label{eq:observer}
  Z(x) \sim \cN\!\big(\bm{\mu}(x),\; \sigma^2(x)\, \mathbf{I}\big),
\end{equation}
whose mean $\bm{\mu}(x)$ and standard deviation $\sigma(x)$ can differ between images.

Second, as with prior work~\cite{compte2000synaptic, wimmer2014bump}, we model memory decay over time as a Brownian motion, where after a delay $\Delta t$, the encoded signal mean $\bm{\mu}(x)$ has moved in a random direction to
\begin{equation}
  \label{eq:diffuse}
  \bm{\mu}(x) + \sqrt{2 D_i \Delta t}\; \bm{\epsilon},
  \qquad \bm{\epsilon} \sim \cN(\mathbf{0}, \mathbf{I}),
\end{equation}
where the diffusion constant $D_i$ is observer specific and controls the magnitude of randomness, effectively influencing how fast an observer $i$ forgets (\Fig{fig:observer}(b)).

Finally, on each trial when presented with an image $y$, the observer draws one sample $\bm{z} \sim Z(y)$ of the current image and finds the shortest distance
to a sample of all earlier images ($\bm{m}_j \sim Z(j)$), $\delta = \min_j \| \bm{z} - \bm{m}_j \|_2$.
The observer reports ``seen'' if $\delta$ is below the decision threshold $\theta_i$, which is observer specific (\Fig{fig:observer}(c)).
Drawing $\bm{z}$, the memories $\{\bm{m}\}$, and their diffusion $\bm{\epsilon}$ many times gives the probability of a ``seen'' response.

\subsection{Fitting the Observer Model Using Behavioral Data}
\label{sec:fitting}

\paragraph{Parameterizing the Perceptual Space.}
We first encode $x$ as $B(x) \in \R^F$ using a frozen, pretrained vision encoder, such as CLIP~\cite{radford2021learning} or one of the ventral-stream areas of CORnet-S~\cite{kubilius2019brain}, a convolutional neural network whose four blocks align with neural recordings in the primate ventral-stream areas V1, V2, V4, and IT.
The encoders are shown to correlate with, and are used as proxies for, various stages in the human visual system.
We then reduce $B(x)$ into an $N=8$-dimensional space using t-SNE~\cite{vandermaaten2008visualizing}, which attempts to keep the nearest neighbors of each image.

\paragraph{Parameterizing the Encoding Variance.}
We let the variance of an image's distribution depend on two image properties, both computed once from frozen encoders:
\begin{equation}
  \label{eq:sigma}
  \sigma(x) = s_0 \exp\big(\gamma\, T(x) - \beta\, A(x)\big),
\end{equation}
where $s_0 > 0$ sets the overall scale.
$T(x)$ measures how distinctive an image is.
We compute it from the distance between the image and its neighbors in the feature space of the vision encoder: $T(x)$ is the mean cosine similarity between the image and its $300$ nearest neighbors, so a common image has a high $T(x)$ and a distinctive image a low $T(x)$.
$A(x)$ is the mean response strength of the IT-layer in CORnet-S model to the image, following the finding that images that evoke a stronger response in the inferotemporal cortex are more memorable~\cite{jaegle2019population}.
\Apx{app:fitting} gives how $T(x)$ and $A(x)$ are computed.

\paragraph{Diffusion over Long Delays.}
Human forgetting is fast at first and slow later~\cite{wixted1991form, rubin1996one}.
We therefore let the diffusion slow down over time: at time $s$ after an image is stored, its memory diffuses at the rate $D_i / (1 + s / t_0)$, so it moves quickly at first and slowly later (\Apx{app:fitting}).
The time scale $t_0$ is shared by all observers and fitted with the other parameters.

\paragraph{Likelihood.}
Given the encoding, the diffusion, and the decision rule, we can now describe the (log) likelihood function:
\begin{equation}
  \label{eq:loglik}
  \log p(\mathcal{O} \mid \Theta) = \sum_{t=1}^{T} \Big[ y_t \log P_t + (1 - y_t) \log\big(1 - P_t\big) \Big],
\end{equation}
where $\mathcal{O}$ represents the responses of all observers on all $T$ trials, $y_t = 1$ if the observer reported ``seen'' on trial $t$, and $P_t$ is the probability of this response predicted by our observer model.
Note that $y_t$ is the response of the \textbf{observer}, not whether the image actually repeated.
The trainable parameter set $\Theta$ is $\{s_0, \gamma, \beta, t_0\}$, which are shared across observers and $\{\theta_i, D_i\}_{i=1}^{I}$, which is specific to each of the $I$ observers.

\paragraph{Model Fitting.}
We estimate $\Theta$ by maximizing the likelihood on the training set of each dataset (\Sect{sec:setup}).
The probability $P_t$ has no closed form, so we resort to a numerical approximation using Monte Carlo (MC) simulations, where we draw $K$ samples of the current image and of all memories and average the decisions:
\begin{equation}
  \label{eq:mc}
  P_t \approx \frac{1}{K} \sum_{k=1}^{K} \mathbf{1}\{ \delta^{(k)} < \theta_i \},
\end{equation}
where $\delta^{(k)}$ is the distance to the nearest memory in simulation run $k$, and $\mathbf{1}\{\cdot\}$ is the indicator function.
The indicator function is not differentiable, so during fitting we use the reparameterization and gradient estimation trick~\cite{kingma2014auto} to make it differentiable.
\Apx{app:fitting} gives more details.

%% file: sections/experiments.tex
\section{Experiment Setup}
\label{sec:setup}

\paragraph{Feature Encoder.}
Our primary choice of encoder $B(\cdot)$ is CLIP~\cite{radford2021learning}: a ViT-L/14 trained on LAION-2B~\cite{schuhmann2022laion} with OpenCLIP~\cite{cherti2023reproducible}, which maps each image to a vector of dimension $F = 768$.
To compare representations along the visual hierarchy, we also use CORnet-S~\cite{kubilius2019brain}.
We spatially pool each channel of each block, obtaining one vector per area of dimension $F = 64, 128, 256, 512$, respectively.
Finally, we use a contrast pyramid~\cite{peli1990contrast}, which approximates the spatial processing by the retina.
We use an implementation ColorVideoVDP~\cite{mantiuk2024colorvideovdp}, pool the feature within each (channel, band) combination ($F = 24$).
The IT-layer response $A(x)$ used to encode the distribution variance in \Equ{eq:sigma} always comes from the IT block of CORnet-S, independent of the feature encoder.
No encoder is fine-tuned.

\paragraph{Datasets.}
We use four continuous recognition datasets that differ in images, participants, and intervals between two appearances of the same image.

\textit{Material}~\cite{wang2026material} contains 121,727 trials from 26 people who viewed 3,511 photographs of materials from the STUFF dataset~\cite{schmidt2025core} and pressed a key when an image repeated.
A target repeats after 41--59 other images.

\textit{MemCat}~\cite{goetschalckx2019memcat} contains 3,011,800 trials from 2,291 people who did the same task with target images from five categories (animals, food, landscapes, sports, and vehicles) mixed with filler images.
Targets repeat after 20--150 other images, within about 3.5 minutes.

\textit{NSD}~\cite{allen2022massive} contains 212,374 trials from 8 people who viewed natural scenes from COCO~\cite{lin2014microsoft} over about 40 scanning sessions and answered old or new on every trial.
Delays range from seconds to about 300 days.

\textit{THINGS}~\cite{kramer2023features} contains 6,062,730 trials from 20,578 people who viewed 26,107 photographs of 1,854 object concepts and pressed a key when an image repeated; delays range from seconds to about 4.5 minutes.
In Material, MemCat, and THINGS, we exclude vigilance repeats.

\paragraph{Evaluation Metrics.}
For every trial, our observer model predicts the probability $P_t$ that the observer reports ``seen'' (\Equ{eq:mc}).
We report three trial-level metrics: the negative log-likelihood (NLL) per trial, which is the quantity that we fit; the binarized accuracy; 
and the agreement score~\cite{zhang2018unreasonable}, $\frac{1}{T} \sum_{t} \big[ y_t P_t + (1 - y_t)(1 - P_t) \big]$, which measures the probability that the model and the observer give the same answer on a trial.

We also calculate per-image statistics, including the average hit rate (HR) of an image across all sessions and all observers, the average false-alarm rate (FAR) of an image, and the memorability score of an image, which is defined as the difference between HR and FAR~\cite{goetschalckx2019memcat}.
As with convention, we report the Spearman's Rank Correlation Coefficient (SRCC) across the image for these three metrics.

\paragraph{Protocol.}
For each dataset, we split the images into two halves at random: a training set and a test set.
All trials whose current image belongs to the test set are used only for evaluation; all observers and all sessions appear in both halves.
The t-SNE embedding, $T(x)$, and $A(x)$ use no behavioral data.

\paragraph{Baselines.}
\textit{Random}: Random predicts ``seen'' with probability 0.5 on every trial, and the base-rate guess predicts each trial by the rate of ``seen'' responses to repeated and to new images in the training set.
\textit{Sequence model}: the LSTM~\cite{hochreiter1997long} reads the sequence of images that an observer has seen, as CLIP features and the time since the previous trial, and predicts the response on each trial; like our model, it is not told whether an image repeats.
\textit{Memorability regression model}: MemNet~\cite{khosla2015understanding}, a convolutional network trained on LaMem, and ResMem~\cite{needell2022embracing}, which builds on ResNet-152~\cite{he2016deep}, regress one memorability score per image.
We train the LSTM on the training set, fine-tune MemNet and ResMem on the memorability scores of the training set.
Full details can be found in \Apx{app:baselines}.

\section{Results}
\label{sec:results}

We first evaluate how well our observer model predicts single recognition decisions (\Sect{sec:baseline}) and image memorability (\Sect{sec:memorability}).
\Sect{sec:ablation} ablates our observer model and shows that our model explains the variation between images.
\Sect{sec:people} shows that our model explains variations across images, and \Sect{sec:encoders} use the model to study which visual representation recognition memory relies on.

\subsection{Observer Model Predicts Per-trial Human Recognition}
\label{sec:baseline}

\begin{table}[t]
\centering
\caption{Comparison of our observer model with the baselines (\Sect{sec:setup}) on the test set.
MemNet and ResMem predict one memorability score per image and no response on single trials, so they have only per-image correlations.
Hit, FA, Mem: correlation with the human per-image hit rate, false-alarm rate, and memorability score.
Params: number of trainable parameters, where $I$ is the number of observers.
Bold: best in each column within a dataset; --: not applicable.
}
\label{tab:baseline}
\footnotesize
\setlength{\tabcolsep}{6pt}
\begin{tabular}{@{}ll ccc ccc r@{}}
\toprule
& & \multicolumn{3}{c}{Per trial} & \multicolumn{3}{c}{Per image ($r$)} \\
\cmidrule(lr){3-5} \cmidrule(l){6-8}
Dataset & Method & NLL\,$\downarrow$ & Acc\,$\uparrow$ & Agr\,$\uparrow$ & Hit\,$\uparrow$ & FA\,$\uparrow$ & Mem\,$\uparrow$ & Params \\
\midrule
Material & Random & 0.693 & 0.500 & 0.500 & -- & -- & -- & 0 \\
         & Base rate & 0.370 & 0.866 & 0.782 & -- & -- & -- & 2 \\
         & LSTM   & 0.495 & 0.804 & 0.702 & 0.23 & 0.33 & 0.02 & 108K \\
         & MemNet & -- & -- & -- & 0.46 & 0.21 & 0.50 & 54.5M \\
         & ResMem & -- & -- & -- & \textbf{0.53} & 0.23 & 0.59 & 69.8M \\
         & Ours   & \textbf{0.347} & \textbf{0.868} & \textbf{0.801} & \textbf{0.53} & \textbf{0.40} & \textbf{0.61} & $4 + 2I$ \\
\addlinespace[3pt]
MemCat   & Random & 0.693 & 0.500 & 0.500 & -- & -- & -- & 0 \\
         & Base rate & 0.328 & 0.881 & 0.806 & -- & -- & -- & 2 \\
         & LSTM   & 0.438 & 0.782 & 0.720 & 0.59 & 0.40 & 0.56 & 50K \\
         & MemNet & -- & -- & -- & 0.73 & 0.27 & 0.77 & 54.5M \\
         & ResMem & -- & -- & -- & \textbf{0.75} & 0.27 & \textbf{0.79} & 69.8M \\
         & Ours   & \textbf{0.305} & \textbf{0.882} & \textbf{0.830} & 0.56 & \textbf{0.41} & 0.53 & $4 + 2I$ \\
\addlinespace[3pt]
NSD      & Random & 0.693 & 0.500 & 0.500 & -- & -- & -- & 0 \\
         & Base rate & 0.595 & 0.689 & 0.592 & -- & -- & -- & 2 \\
         & LSTM   & 0.644 & 0.644 & 0.556 & 0.17 & \textbf{0.61} & $-$0.08 & 50K \\
         & MemNet & -- & -- & -- & $-$0.06 & 0.17 & 0.10 & 54.5M \\
         & ResMem & -- & -- & -- & $-$0.06 & 0.20 & 0.12 & 69.8M \\
         & Ours   & \textbf{0.490} & \textbf{0.770} & \textbf{0.694} & \textbf{0.71} & 0.46 & \textbf{0.59} & $4 + 2I$ \\
\addlinespace[3pt]
THINGS   & Random & 0.693 & 0.500 & 0.500 & -- & -- & -- & 0 \\
         & Base rate & 0.172 & \textbf{0.949} & 0.910 & -- & -- & -- & 2 \\
         & LSTM   & 0.459 & 0.809 & 0.699 & 0.34 & 0.16 & 0.12 & 50K \\
         & MemNet & -- & -- & -- & 0.36 & $-$0.22 & 0.32 & 54.5M \\
         & ResMem & -- & -- & -- & \textbf{0.42} & $-$0.24 & \textbf{0.38} & 69.8M \\
         & Ours   & \textbf{0.167} & 0.947 & \textbf{0.917} & 0.11 & \textbf{0.62} & 0.28 & $4 + 2I$ \\
\bottomrule
\end{tabular}
\end{table}

Our model sets the state of the art on all four datasets for trial-by-trial predictions of human responses in CRM tasks.
The results are shown in \Tbl{tab:baseline}.
It outperforms the strongest baseline, the base-rate guess, although this guess is told whether an image repeats, and it outperforms the LSTM, which reads the same image sequences (24--64\% lower NLL).
Our model needs only four shared parameters and two per observer (\Sect{sec:fitting}), because its structure already specifies how an image is encoded, how its memory fades, and how it is compared with the stored memories.
Memorability networks such as MemNet~\cite{khosla2015understanding} and ResMem~\cite{needell2022embracing} are designed to predict a single memorability score per image, so they cannot predict single trials.

The accuracy of our trial-by-trial prediction is within the inherent variation across human observers.
In particular, we predict each observer’s response on a given trial from the other observers’ responses to the same image.
This is feasible because, recall, that each observer has a model fit with their own decision threshold $\theta_i$ and diffusion constant $D_i$. 
For each held-out observer, we use the models fitted to the other observers to predict their trial-by-trial responses, and compare the resulting accuracy with the human data, which sets the ceiling of inter-observer prediction ceiling.

\Tbl{tab:human} (``Per trial'' columns) shows the results.
Our model comes very close to this ceiling (NLL within 6\% of people on Material, MemCat, and THINGS) set by inherent variation across human observers.
The NLL actually slightly exceeds the ceiling on NSD.
This is because NSD includes only eight observers, so the ceiling is estimated from few responses and is therefore noisy, and our model can exceed it by chance. 

\subsection{Observer Model Predicts Image Memorability}
\label{sec:memorability}
\label{sec:history}

\begin{table}[t]
\centering
\caption{Comparison of our observer model with the human ceiling on the test set.
People shows how well other people predict a response, which serves as the ceiling.
Per trial: we predict each response from the trials of the other observers on the same image, when it is new or when it repeats.
People uses the fraction of these observers who report ``seen'', and Ours uses the mean of our predicted probabilities on their trials.
Per image: we split the observers into two random halves and predict the per-image rates of one half from the trials of the other half.
People uses the other half's responses, and Ours uses our predictions on those trials.
Correlations are averaged over 100 splits, without the Spearman--Brown formula.
Ours is computed in this way to compare it with the human ceiling, so its numbers differ from those in \Tbl{tab:baseline}.
\Apx{app:reference} gives the details.}
\label{tab:human}
\small
\setlength{\tabcolsep}{6pt}
\begin{tabular}{@{}ll ccc ccc@{}}
\toprule
& & \multicolumn{3}{c}{Per trial} & \multicolumn{3}{c}{Per image ($r$)} \\
\cmidrule(lr){3-5} \cmidrule(l){6-8}
Dataset & & NLL\,$\downarrow$ & Acc\,$\uparrow$ & Agr\,$\uparrow$ & Hit\,$\uparrow$ & FA\,$\uparrow$ & Mem\,$\uparrow$ \\
\midrule
Material & Human ceiling & 0.341 & 0.869 & 0.792 & 0.63 & 0.61 & 0.61 \\
         & Ours   & 0.361 & 0.866 & 0.786 & 0.47 & 0.31 & 0.53 \\
\addlinespace[3pt]
MemCat   & Human ceiling & 0.309 & 0.883 & 0.816 & 0.84 & 0.65 & 0.79 \\
         & Ours   & 0.322 & 0.881 & 0.809 & 0.53 & 0.31 & 0.47 \\
\addlinespace[3pt]
NSD      & Human ceiling & 0.595 & 0.693 & 0.620 & 0.54 & 0.38 & 0.29 \\
         & Ours   & 0.587 & 0.688 & 0.609 & 0.60 & 0.35 & 0.43 \\
\addlinespace[3pt]
THINGS   & Human ceiling & 0.175 & 0.948 & 0.908 & 0.33 & 0.14 & 0.19 \\
         & Ours   & 0.177 & 0.949 & 0.896 & 0.10 & 0.02 & $-$0.01 \\
\bottomrule
\end{tabular}
\end{table}

\begin{figure}[t]
\centering
\includegraphics[width=0.6\linewidth]{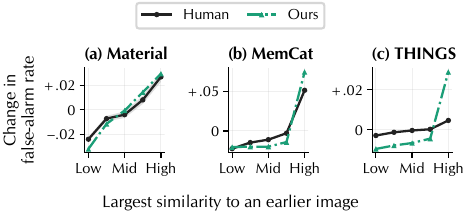}
\caption{Impact of previously seen images on false alarms.
Each curve shows how the false-alarm rate of a new image changes with its largest CLIP similarity to any image that the observer saw earlier.
In people and in our model, a new image is falsely recognized more often after the observer has seen a similar image.
Shaded: the standard error of the human mean across trials, which represents the uncertainty of the human curve.}
\label{fig:history}
\vspace{-8pt}
\end{figure}

Since our model is fit only to single trials, it can directly be used to predict the hit rate, false-alarm rate, and memorability scores of an image, just like how these statistics are obtained on humans.
We report the SRCC of these three metrics in \Tbl{tab:baseline} (the `Per image'' columns).

On most datasets, this score gives the highest false-alarm correlation.
Although our model does slightly worse than the deep-learning methods MemNet and ResMem on MemCat and THINGS, these are pretrained networks with tens of millions of weights, supervised directly on memorability scores and fine-tuned on each dataset, whereas our model has only four shared parameters and two per observer and is never supervised on memorability scores.

Our model also comes very close to the human ceiling on the false-alarm rate, the hit rate, and memorability, as shown in \Tbl{tab:human} (``Per image'' columns).
It performs on par with people on Material and NSD, and slightly worse on MemCat and THINGS.
This is because each observer in MemCat and THINGS (both crowdsourced datasets) completes few trials, so their parameters are estimated less accurately.
So the performances fall slightly below the human ceiling.

A participant who has seen similar images before is more likely to falsely recognize a new image as seen.
Our observer model closely captures this property.
The results are shown in \Fig{fig:history}, where the $x$-axis bins all images based on their similarity to all previously seen images in the CLIP space, and the $y$-axis is the normalized FAR; the normalization is done by subtracting the average FAR across all images.
We see that the more similar an earlier image is to the new image, the more often the participant falsely recognizes it;
the prediction of our observer model rises in the same way, closely aligned with people (0.063 vs.\ 0.077 per $0.1$ on Material).

\subsection{Encoding Variance}
\label{sec:ablation}
\label{sec:encoding}
\label{sec:it}

\begin{figure*}[t]
\centering
\includegraphics[width=0.98\textwidth]{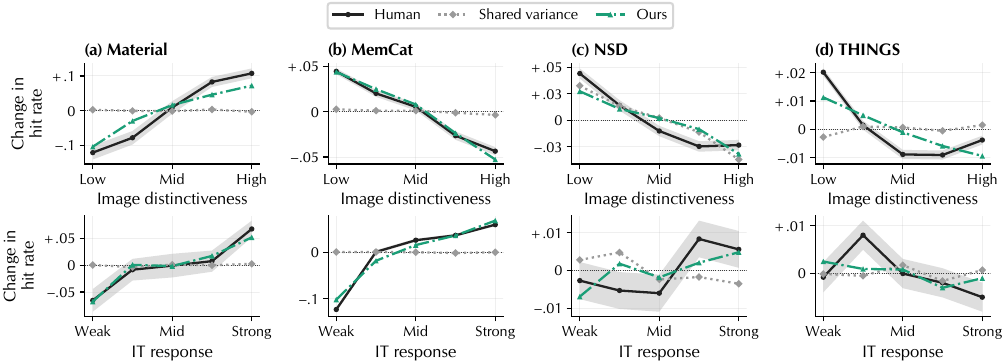}
\caption{Impact of image distinctiveness and IT-layer response on the hit rate.
Each panel shows how the per-image hit rate changes with image distinctiveness (top) or with the IT-layer response (bottom), for people, our model, and our model with one encoding variance shared by all images (Shared variance).
Our model follows the human trends on all four datasets.
The shaded band shows the standard error of the human mean across images, which represents the uncertainty of the human curve.
The $y$-axis shows the normalized hit rate, where the normalization is done by subtracting the average hit rate across all images.
}
\label{fig:encoding}
\label{fig:it}
\end{figure*}

\begin{table}[t]
\centering
\caption{Contribution of each part of our observer model on the test set.
Each row changes one part of our model and refits the model.
The rows change how the variance is encoded (\Equ{eq:sigma}) or how memories are forgotten.
The last row is our model.}
\label{tab:ablation}
\small
\setlength{\tabcolsep}{4pt}
\begin{tabular}{@{}l cc cc cc cc@{}}
\toprule
& \multicolumn{2}{c}{Material} & \multicolumn{2}{c}{MemCat} & \multicolumn{2}{c}{NSD} & \multicolumn{2}{c}{THINGS} \\
\cmidrule(lr){2-3} \cmidrule(lr){4-5} \cmidrule(lr){6-7} \cmidrule(l){8-9}
& NLL\,$\downarrow$ & Mem\,$\uparrow$ & NLL\,$\downarrow$ & Mem\,$\uparrow$ & NLL\,$\downarrow$ & Mem\,$\uparrow$ & NLL\,$\downarrow$ & Mem\,$\uparrow$ \\
\midrule
\multicolumn{9}{@{}l}{\textit{Encoding variance}} \\
Shared & 0.3483 & 0.22 & 0.3108 & 0.07 & 0.4899 & 0.59 & 0.1666 & 0.26 \\
Distinct. only & 0.3465 & 0.55 & 0.3096 & 0.10 & 0.4898 & 0.60 & 0.1666 & 0.28 \\
IT only & 0.3488 & 0.41 & 0.3062 & 0.55 & 0.4898 & 0.59 & 0.1670 & 0.26 \\
\midrule
\multicolumn{9}{@{}l}{\textit{Forgetting}} \\
Const. diffusion & 0.3473 & 0.61 & 0.3064 & 0.53 & 0.5093 & 0.54 & 0.1673 & 0.28 \\
Classical decay~\cite{wixted1991form} & 0.3510 & 0.55 & 0.3060 & 0.53 & 0.4948 & 0.55 & 0.1682 & 0.28 \\
\midrule
Ours & 0.3467 & 0.61 & 0.3051 & 0.53 & 0.4899 & 0.59 & 0.1673 & 0.28 \\
\bottomrule
\end{tabular}
\end{table}

\begin{figure*}[t]
\centering
\includegraphics[width=0.98\textwidth]{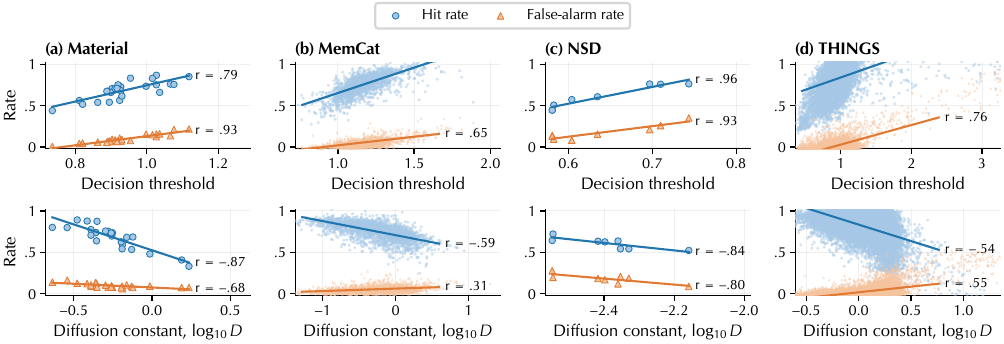}
\caption{Per-observer parameters and differences between people.
Each panel plots the fitted decision threshold (top) or diffusion constant (bottom) of each observer against the observer's hit and false-alarm rates on the test set.
A larger threshold raises both rates, and a larger diffusion constant lowers both rates on Material and NSD.
Each panel removes the linear effect of the other parameter from both axes, so $r$ is a partial correlation.}
\label{fig:people}
\end{figure*}

\begin{figure}[t]
\centering
\includegraphics[width=0.6\linewidth]{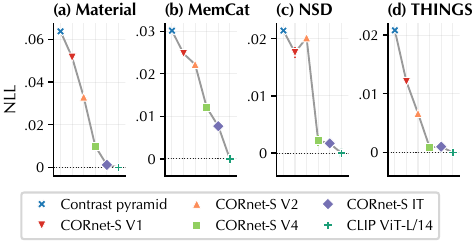}
\caption{Impact of the visual hierarchy stage on predicting recognition.
We show the test-set NLL of our observer model built on encoders from successive stages of the visual hierarchy.
Each panel subtracts the NLL of the model built on CLIP, so CLIP lies at 0.
On all four datasets, the NLL falls from the contrast pyramid to IT, and CLIP gives the lowest NLL.
Error bars: SD over three fitting seeds.}
\label{fig:encoders}
\end{figure}

The encoding variance, set by the distinctiveness of an image and its IT-layer response, explains much of why images differ in memorability.
The results are shown in \Fig{fig:encoding}, where the $x$-axis group images based on their distinctiveness image (top row) and IT-layer reponse strength (bottom row) and the $y$-axis shows the corresponding hit rate normalized by subtracting the average hit rate across all images.
The black lines represent the human data, the gray dotted lines represent the model prediction if distribution variance is encoded independent of the image distinctiveness or IT response strength, and the green lines represent our current model prediction where variance is allowed to be influenced by distinctiveness or IT response strength.

We find that image distinctiveness and IT strength significantly influence the hit rate.
For instance, on the Material dataset, the hit rate is positively correlated with these two metrics, and the hit rate negatively correlates with image distinctiveness on the MemCat and NSD dataset.
Importantly, sharing the distribution variance across images completely fails to capture these trends.

\Tbl{tab:ablation} ablates the influence of variance encoding, where we compare the prediction accuracies (NLL and memorability correlation) when the model uses a shared variance, or when the variance is influenced by image distinctiveness or IT strength alone;
the last row represents our final, comprehensive model.
With one variance shared by all images, the memorability correlation drops on Material and MemCat (0.22 vs.\ 0.61 and 0.07 vs.\ 0.53) and the NLL rises on both, whereas encoding variance using either factor alone recovers the correlation on only one of the two datasets.

\subsection{Decision Threshold and Diffusion Constant Capture Differences Between People}
\label{sec:people}
\label{sec:time}

We find that the two per-observer parameters, decision threshold $\theta_i$ and diffusion constant $D_i$, capture the variation in human observers.
\Fig{fig:people} plots the fitted $\theta_i$ and $D_i$ of each observer against the observer's hit and false-alarm rates.
A larger threshold makes an observer report more images as seen, so $\theta_i$ rises with both rates, and a larger diffusion constant spreads the memories farther away, so $D_i$ falls with both rates on Material and NSD.
Both trends follow the intuition behind our model.

How fast do memories fade?
In our model, the diffusion rate slows down over time (\Sect{sec:method}).
\Tbl{tab:ablation} (Forgetting) compares it with a constant rate and with a classical decay~\cite{wixted1991form}, where a memory does not drift, but the decision threshold for an older memory is smaller (\Apx{app:ablation}).
On NSD, whose delays reach about 300 days, the slowing rate fits best (NLL 0.490 vs.\ 0.509 for a constant rate), because a constant rate cannot forget quickly within a session and slowly across months at the same time.
Classical decay~\cite{wixted1991form} fits worse on Material and NSD (NLL 0.351 vs.\ 0.347 and 0.495 vs.\ 0.490).
Thus, forgetting in recognition memory is fast at first and slow later~\cite{wixted1991form}, and diffusion describes it better than classical decay~\cite{wixted1991form}.

\subsection{Semantic Representations Best Explain Recognition Memory}
\label{sec:encoders}

The higher the stage of the visual hierarchy, the better our model predicts recognition memory.
In \Fig{fig:encoders}, we build the perceptual space from successive stages of the visual hierarchy, from the contrast pyramid~\cite{peli1990contrast} through the four CORnet-S areas~\cite{kubilius2019brain} to CLIP~\cite{radford2021learning}, and keep the rest of the model the same.
On all four datasets, the NLL falls from the contrast pyramid to IT, and CLIP gives the lowest NLL.
This suggests that recognition memory relies on high-level semantic representations of images rather than on their low-level appearance.
DINOv2~\cite{oquab2024dinov2}, SigLIP~\cite{zhai2023sigmoid}, and DreamSim~\cite{fu2023dreamsim} give results close to those of CLIP, so the result does not depend on the choice of the semantic-level vision encoder (\Apx{app:encoders}).

%% file: sections/discussion.tex
\section{Discussion}
\label{sec:discussion}

\paragraph{Interpreting Memorability Scores.}
Memorability is defined as one score per image, which averages over observers and over the images shown before~\cite{isola2011makes, bainbridge2013intrinsic}.
Our model makes this average explicit: each observer, after each history, has a probability of recognizing an image, and the score averages these probabilities.
Two things follow.
First, the memorability of an image is not fixed by the image: it changes with the observers and with what they have seen before (\Sect{sec:history}), as \citet{bylinskii2015intrinsic} found when the same images were shown among different sets of images.
Second, a measured score is a noisy estimate of this average, because each observer responds to each image once, and many images in large datasets are seen by few observers or by one.
Comparisons with the agreement between groups of people should therefore be read as comparisons with the reliability of the data, not with the limit of what a model of single decisions can predict.

\paragraph{Connection to Global Matching Models.}
Our observer model can be read as a detection model~\cite{green1966signal} whose familiarity value is the distance to the nearest memory, computed in a perceptual space built from images.
Global matching models~\cite{gillund1984retrieval, hintzman1988judgments, shiffrin1997model} instead sum the similarity to all memories.

\paragraph{Potential Applications.}
Our observer model can be used beyond predicting memorability.
In neuroscience, single neurons recorded in the medial temporal lobe of epilepsy patients respond differently to new and familiar images~\cite{rutishauser2015representation}, and the same model could be fit to these neural responses as well as to key presses.
Such a model could also guide hippocampal prostheses, which already improve recall in some patients~\cite{hampson2018developing}.
In clinical research, memorability has been used to choose images of a memory test for patients with cognitive decline~\cite{bainbridge2019memorability}, and per-observer parameters of our model could further show whether a patient's memory differs from that of healthy people in memory decay and in the decision.
In computer vision and graphics, predicted memorability is used to compress images~\cite{khanna2019memorability}, to design visualizations~\cite{borkin2013makes}, and to rank videos for media~\cite{cohendet2019videomem}.
Our model could improve image memorability.

\paragraph{Limitations.}
The IT-layer responses come from a model of the ventral visual stream rather than neural recordings, and our model focuses on recognition decisions without predicting response times.

%% file: sections/conclusion.tex
\section{Conclusion}

This paper proposes a framework to model human recognition memory for natural images at the level of single-trial decisions, which lets us capture individual variability at each stage of the memory process. We show that encoding each image as a distribution, whose variance is larger for common images and for images with a weak IT-layer response, explains why some images are harder to recognize than others, and that a threshold and a diffusion constant per person capture how people differ.
We also show that what an observer has seen before changes which new images the observer falsely recognizes, so the memorability of an image is an average of single decisions over observers and histories rather than a property of the image alone.
Among representations from the retina to semantic embeddings, semantic representations best explain which images people remember.
By providing a principled bridge between single recognition decisions and image memorability, we open a new window into how people remember what they see.

%% file: sections/appendix.tex
\section{Model Fitting}
\label{app:fitting}

\paragraph{Perceptual Space.}
We use the image embedding of OpenCLIP ViT-L/14~\cite{radford2021learning}, with each image resized whole to the input size of the encoder.
We run exact t-SNE~\cite{vandermaaten2008visualizing} once on the features of all images of a dataset, with perplexity 30 and an initialization from PCA.
The embedding is not changed during fitting.

\paragraph{Image Properties.}
We compute $T(x)$ and $A(x)$ in \Equ{eq:sigma} once, before fitting.
$A(x)$ is the logarithm of the output of the IT block of CORnet-S, averaged over all locations and channels, with each image resized to 256 pixels.
We standardize $T(x)$ and $A(x)$ with their mean and standard deviation over the images of the training set.

\paragraph{Candidate Memories.}
A sequence can hold hundreds of earlier memories, and simulating all of them on every trial is costly.
For each trial, we keep the memory of the latest presentation of the current image, if the image appeared before, and the $C = 50$ other memories whose means lie nearest to the current image.
Choosing these memories is a discrete step, so we make the choice without gradients.
The chosen memories are then simulated with gradients, so the fit still moves their positions.

\paragraph{Monte Carlo Sampling.}
On every trial, we draw $K = 32$ samples of each candidate memory (\Equ{eq:observer} and \Equ{eq:diffuse})~\cite{metropolis1949monte, mohamed2020monte, zhao2026gazeflow}.
We write each sample as its mean plus a scaled standard Gaussian, so that the samples are differentiable with respect to $s_0$, $\gamma$, $\beta$, $t_0$, and $D_i$~\cite{zhao2026multidimensional}.
We estimate $P_t = (k_t + 0.5) / (K + 1)$, where $k_t$ is the number of simulation runs that report ``seen''.
The added half count keeps $P_t$ away from 0 and 1, so that the NLL of every trial is bounded.

\paragraph{Diffusion over Long Delays.}
With the diffusion rate $D_i / (1 + s / t_0)$, the variance added after a delay $\Delta t$ is
\begin{equation}
  \label{eq:diffusion-age}
  2 \int_0^{\Delta t} \frac{D_i}{1 + s / t_0}\, ds = 2 D_i t_0 \log\!\big(1 + \Delta t / t_0\big),
\end{equation}
so the variance $2 D_i \Delta t$ in \Equ{eq:diffuse} becomes $2 D_i t_0 \log(1 + \Delta t / t_0)$, and the two agree for delays much shorter than $t_0$.

\paragraph{Differentiable.}
The indicator in \Equ{eq:mc} has zero gradient almost everywhere, so during fitting we replace it in every simulation run with
\begin{equation}
  \label{eq:smooth}
  \mathbf{1}\{ \delta < \theta_i \} \approx \mathrm{sigmoid}\Big( \frac{\theta_i - \delta}{\kappa\, \sigma(x)} \Big),
\end{equation}
whose width is a fraction $\kappa = 0.03$ of the standard deviation $\sigma(x)$ of the current image $x$, so it adds no parameter.
We average the sigmoids in place of the decisions~\cite{zhao2026multidimensional}.
All reported results use the indicator.

\paragraph{Optimization.}
The trainable parameters are $s_0$, $\gamma$, $\beta$, and $t_0$, shared by all observers, and $\theta_i$ and $D_i$ of each observer.
We minimize the NLL with Adam~\cite{kingma2015adam}, with learning rate $0.03$, cosine annealing~\cite{loshchilov2017sgdr}, and batch size $256$.
We run $3{,}200$ steps on Material and NSD and $12{,}800$ steps on MemCat and THINGS, which have many more observers.
We fit every model with three random seeds and report the mean.

\section{Datasets and Baselines}
\label{app:datasets}
\label{app:baselines}

\paragraph{Datasets.}
In \textit{Material}, each observer completes three sessions of about 1,550 trials, and no image appears in two sessions.
Vigilance images repeat after 1--4 other images.
In \textit{MemCat}, each observer completes blocks of 200 trials, and no image repeats across blocks.
Vigilance images repeat after 1--7 other images.
In \textit{NSD}, there are no vigilance images; an image is shown up to three times, and every later presentation is a repeated trial.
In \textit{THINGS}, vigilance images repeat after at most 9 other images.
In all datasets, the vigilance repeats are excluded from fitting and evaluation but stay in the history of the observer, because the observer still saw them.

\paragraph{MemNet and ResMem.}
MemNet~\cite{khosla2015understanding} is released only as a Caffe model trained on LaMem.
We parse the Caffe weights into PyTorch and check the port on the first LaMem test split, where it reaches a Spearman correlation of 0.638, against 0.64 reported by the authors.
The released ResMem~\cite{needell2022embracing} is trained on LaMem and MemCat, so it has seen the MemCat images of our test set.
We therefore retrain ResMem on the first LaMem training split, with the released architecture and input.
The ImageNet ResNet-152 inside ResMem stays frozen, and the other layers are trained from scratch with the mean squared error, Adam with learning rate $10^{-4}$, and batch size 64, stopping early on the first LaMem validation split.
The retrained network reaches a Spearman correlation of 0.666 on the first LaMem test split.

We fine-tune both networks on each dataset in the same way.
The target of an image is its memorability score in the training set, computed from all observers, for every fitting image with at least one repeated and one new trial.
We keep the convolutional layers and the ResNet-152 frozen and continue training the fully connected layers, with the mean squared error, Adam with learning rate $10^{-4}$ and weight decay $10^{-4}$, batch size 32, and one random view of each image per step.
We stop early on 20\% of the fitting images (patience 15), average the predictions over 10 views at test time, and report the mean over five seeds.

Both networks are pretrained on LaMem, which may share images with our test sets.
We embed every LaMem image with CLIP and count the images of the test set whose largest cosine similarity to a LaMem image exceeds 0.95: none of the 1,755 images of Material, 146 of 10,000 on MemCat, 1,649 of 35,306 on NSD, and 9 of 13,053 on THINGS.
Removing these images changes every per-image correlation of MemNet and ResMem in \Tbl{tab:baseline} by less than 0.01.

\paragraph{LSTM.}
The input of each trial is the CLIP feature of the image, standardized and mapped by a linear layer with a tanh to 32 dimensions, together with the logarithm of the time since the previous trial.
A one-layer LSTM reads the trials in order, and a linear layer on its hidden state gives the probability of a ``seen'' response.
A sequence is one session of an observer on Material, one block on MemCat, and all trials of an observer on NSD and THINGS.
The loss is the NLL on the trials of the training set; the trials of the test set stay in the sequence as input.
We train two configurations with AdamW: 128 hidden units (108K parameters), and 64 hidden units with dropout 0.3 (50K parameters).
We hold out 10\% of the training images for validation and keep the configuration with the lowest validation NLL.

\section{Comparison with Human Observers}
\label{app:reference}

In \Tbl{tab:human}, both people and our model predict the responses of some observers from the trials of other observers, so neither has seen the responses that it predicts~\cite{zhao2026multidimensional}.

\paragraph{Per Trial.}
For each test trial of an observer, we take the test trials of all other observers on the same image in the same condition, new or repeated.
People predicts the trial by the fraction of these trials with a ``seen'' response, $(n_{\text{seen}} + 0.5) / (n + 1)$, where $n$ is the number of these trials.
Ours predicts it by the mean of our predicted probabilities on the same trials.
Neither uses the test responses of the observer who is predicted.
On Material, MemCat, and THINGS, every test trial has trials of other observers.
On NSD, few images are shown to more than one observer, so only 10,614 test trials (10\%) have them, with at most seven other observers each, and the NSD row of \Tbl{tab:human} uses these trials.

\paragraph{Per Image.}
For each of 100 random splits, we divide the observers into two halves, A and B, and compute the per-image rates of each half on the test set that both halves saw.
People correlates the rates of half A with those of half B.
Ours correlates the mean of our predicted probabilities on the trials of half A with the rates of half B.
We average the correlations over the splits.
On NSD, the comparison uses the 506 images of the test set that all observers saw.

\section{Ablation Variants}
\label{app:ablation}

We define here the variants of \Tbl{tab:ablation} and two more variants, PCA and power-law slowing, whose results are in \Tbl{tab:variants}.
Each variant changes one part of our model and refits it with three seeds.

\begin{table}[H]
\centering
\caption{NLL of two more variants of our observer model on the test set, mean over three seeds.
PCA changes the perceptual space, and power-law slowing changes the diffusion rate.}
\label{tab:variants}
\small
\setlength{\tabcolsep}{4pt}
\begin{tabular}{@{}l cccc@{}}
\toprule
& Material & MemCat & NSD & THINGS \\
\midrule
PCA               & 0.3560 & 0.3175 & 0.5016 & 0.1661 \\
Power-law slowing & 0.3455 & 0.3051 & 0.4926 & 0.1671 \\
\midrule
Ours              & 0.3467 & 0.3051 & 0.4899 & 0.1673 \\
\bottomrule
\end{tabular}
\end{table}

\paragraph{Perceptual Space.}
PCA replaces t-SNE with the first $N = 8$ principal components of the CLIP features of the training set.
t-SNE gives a lower NLL on Material, MemCat, and NSD, by 0.009 to 0.012, and the two differ by about 0.001 on THINGS (\Tbl{tab:variants}).

\paragraph{Encoding Variance.}
Shared uses one standard deviation $s_0$ for all images ($\gamma = \beta = 0$ in \Equ{eq:sigma}).
Distinctiveness only sets $\beta = 0$, and IT only sets $\gamma = 0$.

\paragraph{Forgetting.}
Constant diffusion keeps the rate at $D_i$ at all times.
Power-law slowing uses the rate $D_i (1 + s / t_0)^{-b}$ and fits the exponent $b$, which is 1 in our model.
The NLL stays within 0.003 of our model on all four datasets (\Tbl{tab:variants}), so we keep $b = 1$.
Decay keeps every memory where it was stored and instead penalizes old memories: the distance to memory $j$ becomes $\| \bm{z} - \bm{m}_j \|_2 + a_i \log(1 + \Delta t_j / t_0)$, where $a_i$ is a per-observer decay rate that takes the place of $D_i$.
An older memory must therefore lie closer to the current image to be recognized, which acts as a smaller decision threshold for older memories.

\section{Results on Other Encoders}
\label{app:encoders}

We replace CLIP with one of three other encoders: DINOv2~\cite{oquab2024dinov2}, SigLIP~\cite{zhai2023sigmoid}, and DreamSim~\cite{fu2023dreamsim}.
Only the perceptual space changes: $T(x)$ is computed in the feature space of each encoder, $A(x)$ is the same for all encoders, and $N = 8$.

\Tbl{tab:encoders} gives the results on Material and MemCat.
All three encoders give an NLL within 0.01 of CLIP, and their correlations with the human memorability scores are within 0.04 of CLIP.
The result of \Sect{sec:encoders} therefore does not depend on the choice of the semantic-level vision encoder.

\begin{table}[H]
\centering
\caption{Our observer model with different encoders on the test set, mean over three seeds.
Hit, FA, Mem: correlation with the human per-image hit rate, false-alarm rate, and memorability score.}
\label{tab:encoders}
\setlength{\tabcolsep}{5pt}
\begin{tabular}{@{}ll cccc@{}}
\toprule
Dataset & Encoder & NLL\,$\downarrow$ & Hit\,$\uparrow$ & FA\,$\uparrow$ & Mem\,$\uparrow$ \\
\midrule
Material & CLIP     & 0.3467 & 0.53 & 0.40 & 0.61 \\
         & DINOv2   & 0.3505 & 0.47 & 0.40 & 0.57 \\
         & SigLIP   & 0.3455 & 0.50 & 0.40 & 0.61 \\
         & DreamSim & 0.3382 & 0.40 & 0.51 & 0.59 \\
\midrule
MemCat   & CLIP     & 0.3051 & 0.56 & 0.41 & 0.53 \\
         & DINOv2   & 0.3068 & 0.53 & 0.37 & 0.56 \\
         & SigLIP   & 0.3097 & 0.56 & 0.33 & 0.54 \\
         & DreamSim & 0.3062 & 0.54 & 0.44 & 0.54 \\
\bottomrule
\end{tabular}
\end{table}

\section{Results on Individual Differences}
\label{app:people}

The introduction states how much of the differences in performance between observers each per-observer parameter explains.
We measure the performance of each observer by the hit rate minus the false-alarm rate on the test set, and regress it on the fitted $\theta_i$ and $\log D_i$.
The two parameters are correlated, so their individual $R^2$ values are not additive. 
We use Shapley values to divide the joint $R^2$ between the two parameters.

\Tbl{tab:people} gives the shares.
On all three datasets, the fitted diffusion constant explains more of the differences between observers than the fitted decision threshold: 59\% against 13\% on Material, 34\% against 22\% on MemCat, and 35\% against 5\% on THINGS.
We leave out NSD, which has only eight observers.

\begin{table}[H]
\centering
\caption{Share of the differences in performance between observers that each per-observer parameter explains.
Performance is the hit rate minus the false-alarm rate on the test set.
Total: $R^2$ of both parameters together.}
\label{tab:people}
\setlength{\tabcolsep}{5pt}
\begin{tabular}{@{}l ccc@{}}
\toprule
Dataset & $D_i$ & $\theta_i$ & Total \\
\midrule
Material & 59\% & 13\% & 72\% \\
MemCat   & 34\% & 22\% & 56\% \\
THINGS   & 35\% & 5\%  & 40\% \\
\bottomrule
\end{tabular}
\end{table}